\documentclass[singlecolumn]{IEEEtran}          
\usepackage{graphicx}
\usepackage[caption=false]{subfig}
\usepackage{epstopdf}
\usepackage{comment}
\usepackage{color}
\usepackage[colorinlistoftodos,prependcaption,textsize=tiny]{todonotes}
\usepackage{stackengine}
\usepackage{comment}
\usepackage{amsmath}

\newcommand{\rcite}[1]{\textcolor{red}{\cite{}}}
\usepackage{caption}

\begin{document}
	\title{Training Deep Face Recognition Systems\\ with Synthetic Data}
	\author{Adam Kortylewski, \and
		Andreas Schneider,	\and
		Thomas Gerig,	 \and
		Bernhard Egger,	\and\\
		Andreas Morel-Forster, \and
		Thomas Vetter\\\vspace{.1cm}
	    Department of Mathematics and Computer Science\\
	    University of Basel
    }

	\date{}
	\maketitle
	\begin{abstract}
    Recent advances in deep learning have significantly increased the performance of face recognition systems. The performance and reliability of these models depend heavily on the amount and quality of the training data. However, the collection of annotated large datasets does not scale well and the control over the quality of the data decreases with the size of the dataset. 
    In this work, we explore how synthetically generated data can be used to decrease the number of real-world images needed for training deep face recognition systems. In particular, we make use of a 3D morphable face model for the generation of images with arbitrary amounts of facial identities and with full control over image variations, such as pose, illumination, and background. In our experiments with an off-the-shelf face recognition software we observe the following phenomena: 1) The amount of real training data needed to train competitive deep face recognition systems can be reduced significantly. 2) Combining large-scale real-world data with synthetic data leads to an increased performance. 3) Models trained only on synthetic data with strong variations in pose, illumination, and background perform very well across different datasets even without dataset adaptation. 4) The real-to-virtual performance gap can be closed when using synthetic data for pre-training, followed by fine-tuning with real-world images. 5) There are no observable negative effects of pre-training with synthetic data. Thus, any face recognition system in our experiments benefits from using synthetic face images. The synthetic data generator, as well as all experiments, are publicly available.    
	\end{abstract}

    \section{Introduction}

    Face recognition is a major field in computer vision. In recent years, the community shifted from engineering features by hand to using deep learning approaches. As a result of this paradigm shift, the general face recognition performance increased massively. Among others, work by Facebook \cite{taigman2014deepface} and Google \cite{schroff2015facenet} demonstrated that a training with large-scale datasets leads to a close-to-maximum performance on standard benchmarks, such as LFW \cite{lfw}, YTF\cite{ytf} or IJB-A\cite{ijba}. Hence, it became apparent that deep learning approaches strongly rely on large and complex training sets to generalize well in unconstrained settings. However, the common approach of collecting and maintaining large-scale datasets has two severe disadvantages, which we aim to overcome in this work:
    
    1) It is practically unfeasible to collect large-scale training datasets for advanced face recognition tasks, such as e.g. disguised or occluded identities in difficult illumination conditions or strong pose variations. However, with the introduction of more complex benchmark datasets such as MegaFace \cite{kemelmacher2016megaface}, IJB-B \cite{ijbb}, IJB-C\cite{ijbc} and the recently proposed disguised face recognition challenge \cite{dfw}, it becomes apparent that in the future more complex and thus rarer data is needed for training. This data is difficult to obtain, since it is time-intensive to acquire and annotate. We aim to overcome this problem by using a parametric face image generator to synthesize arbitrary amounts of face images.

    2) Large data sets are not just difficult to collect but are also hard to control in terms of dataset variability. It is not scalable to annotate all the variations of interest such as e.g. pose, expression or illumination. In addition, a labor-intensive ground truth annotation process is also prone to errors. However, controlling the distribution of these variables is of critical importance for face-recognition applications since these are often deployed in security systems. A bias or a blind spot in these systems could have far-reaching implications. Synthetic data offers full control and detailed reliable annotations for free. In our experiments, we show that synthetic face images are complementary to biased real-world datasets resulting in an increased recognition performance across several recognition benchmarks.
    
    Observing these challenges and the fact that statistical models of faces in 3D can be combined with computer graphics to synthesize face images motivates the major research question that guides this work: 
    
	\begin{center}
	    \textit{Can we use synthetically generated data to support the training and reliability of face recognition systems?}
	\end{center}  
    
    3D Morphable Face Models (3DMMs) \cite{blanz1999morphable} are an active field of research since decades. 
    Rendering synthetic 2D images of faces is a basic capability of 3DMMs.  We will leverage this capability in order to generate large scale datasets, which will then be used to train deep face recognition systems. 
    The parametric nature of the 3DMM thereby provides full control over the facial identities in the terms of shape and albedo texture in an image, but also over nuisance parameters such as pose, illumination and the facial expression. In addition, 3DMMs enable us to create possibly infinite data sets in arbitrary depth (number of samples per identity) and width (number of identities). The synthetic images can easily be released publicly for research purposes as the 3DMM is publicly available. However, the major advantage of our approach is that it opens an alternative approach to the acquisition and labelling of ground truth data for new face recognition problems. Namely, the possibility to model the problem generatively, which is much simpler for many facial appearance variations e.g. due to changing illumination conditions or head pose variations. A critical part of the research question is whether deep face recognition systems can transfer the knowledge acquired from synthetic data and combine it with real data. Because although computer graphics and 3D facial modelling have made great progress, some variations of face images currently cannot be represented realistically in a parametric model, such as the skin texture and wrinkles or facial occluders such as beards or scarfs. 
    
    In this work, we analyze this research question in in detail. Thereby, we use synthetically generated data to train a vanilla OpenFace \cite{openface} implementation of the FaceNet architecture \cite{schroff2015facenet}. 
    
    Our analysis documents that:
    \begin{enumerate}
        \item A significant real-to-virtual performance gap exists between neural networks trained on synthetic and real-world data, which can be closed by fine-tuning with real-world data.
        \item Using synthetic data, we can either reduce the amount of real data needed to achieve competitive face recognition performance or combine it with large-scale real datasets to increase the recognition performance significantly across several standard benchmarks without database adaptation.
        \item Already good face recognition performances are achieved by using purely synthetic data, which is generated from just 200 real 3D scans.
        \item We did not observe any negative effects when pre-training with synthetic data, thus the performance gains come for free
    \end{enumerate}
    The paper is structured as follows: We discuss related work in Section~\ref{sec:related} and introduce our face image generator in Section~\ref{sec:generator}. We study in detail how synthetic data can support the training and reliability of face recognition systems in Section~\ref{sec:exp}. We conclude our work and discuss caveats in Section~\ref{sec:conclusion}.
    
    \section{Related Work}
    \label{sec:related}     
    \textbf{Deep Face Recognition.} The performance of face recognition systems has significantly increased with the introduction of deep convolutional neural networks. Taigman et al. \cite{taigman2014deepface} used four million images to train a siamese network architecture, where the distance between feature embeddings of images from the same person are minimized. Further, architectural changes have since then steadily increased face recognition performances such as DeepID 1-3 \cite{sun2014deep2,sun2014deep1,sun2015deepid3} or the pose-aware model of Masi et al. \cite{masi2016pose}. Googles' FaceNet \cite{schroff2015facenet} combined a dataset of 200 million images from eight million identities with an innovative triplet-loss to achieve a novel state of the art performance at that time. Later, Parkhi et al. \cite{parkhi2015deep} have shown that dataset sizes can be cut down by a factor of ten with an adjusted network architecture. 
    We also use the triplet-loss optimization procedure throughout our experiments together with the FaceNet architecture as presented in \cite{schroff2015facenet}.
    
    \textbf{Face Recognition with augmented data.} Although, changes of the network architecture have played an important role, 
    having access to massive training datasets has been the critical component for deep face recognition systems to achieve ever higher face recognition rates. However, the collection, annotation, and publication of millions of face images have proven to be practically unfeasible. Therefore, a major branch of research has evolved around techniques to augment available data in order to increase dataset sizes artificially. Rudimentary approaches to data augmentation are geometric transformations such as mirroring \cite{chatfield2014return,yang2015mirror}, translational shift \cite{levi2015age} and rotation \cite{wolf2011effective,chen2012dictionary}. 
    Patel et al. \cite{patel2011illumination} perform face image relighting in order to be robust to strong illumination changes between images of the same person. Hu et al. \cite{hu2018frankenstein} compose novel face images by blending parts of different donor face images into a novel image.
    Masi et al. \cite{masi2016we} use domain-specific data augmentation by augmenting the Casia dataset \cite{casia} in 3D. They align a 3D face shape to landmarks in each face image and subsequently change the face in the image in terms of shape deformation, 3D pose, and expression neutralization. This augmentation process, however, is very generic and does not follow any statistical distribution. 
    Furthermore, the illumination of the images cannot be changed, thus relying on the illumination distribution of the Casia dataset. In this work, we follow a different approach by studying the effect of using fully synthetic images which were generated with statistical 3DMMs, a statistical illumination model, and computer graphics.
    
    \textbf{Deep learning and fully synthetic data.} 
    Fully synthetic datasets which are generated with computer graphics have been widely used for the evaluation and training of computer vision tasks such as optical flow \cite{butler2012naturalistic}, autonomous driving systems \cite{chen2015deepdriving}, object detection \cite{gupta2014learning}, pose estimation \cite{shotton2011real,park2015articulated,ionescu2014human3} or text detection \cite{gupta2016synthetic}. Recently, Qiu and Yuille \cite{qiu2017unrealcv} developed UnrealCV, a computer graphics engine for the \textit{diagnosis} of computer vision algorithms at scene analysis. Following this approach, we recently proposed to use synthetically generated face images for analyzing the generalization performance of different neural network architectures at face recognition in the virtual domain \cite{kortylewski2017empirically}. Gaidon et al. presented Virtual KITTI \cite{gaidon2016virtual}, where they use synthetically generated data to pre-train a deep convolutional neural network at the task of object detection, tracking and scene segmentation in the context of automated driving systems. They show that Deep Learning systems behave similarly when trained in the synthetic domain and evaluated in the real domain and vice-versa. Concerning facial analysis, Abbasnejad et al. \cite{abbasnejad2017using} train a deep convolutional neural network for expression analysis on synthetic data and achieve state of the art results in action unit classification on real data. Curiously, despite the widespread use of synthetic data for deep learning, we are not aware of any work studying how synthetic data can be leveraged to support face recognition in real-world images.
    
    In this work, we use synthetic data generated from sampling a statistical model of 3D face shape, texture, expression and illumination in order to generate synthetic face images, which we use to train deep convolutional neural networks for face recognition.

	\section{Face Image Generator}
	\label{sec:generator}
	
    \begin{figure}
    \centering
    \stackunder[2pt]{\includegraphics[width=.15\textwidth]{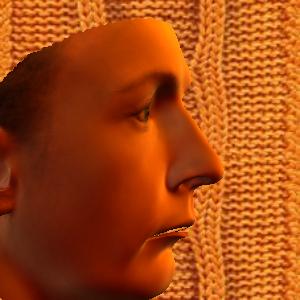}}{}
    \stackunder[2pt]{\includegraphics[width=.15\textwidth]{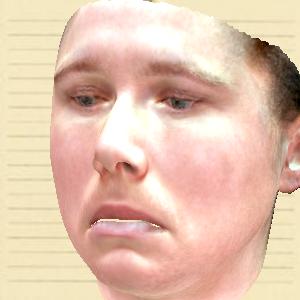}}{}\\
    \stackunder[5pt]{\includegraphics[width=.15\textwidth]{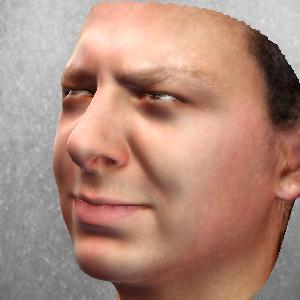}}{}
    \stackunder[5pt]{\includegraphics[width=.15\textwidth]{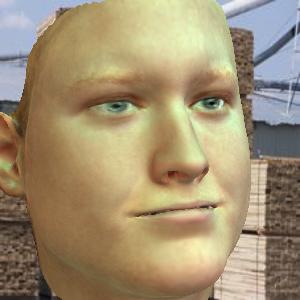}}{}
    \caption{Example renderings generated by our parametric face image generator. The facial appearance in the images varies in terms of identity, head pose, expression and illumination changes.}
    \label{fig:exampleSynth}
    \end{figure}
	
	We propose to synthesize face images by sampling from a statistical 3D Morphable Model \cite{blanz1999morphable} of face shape, color and expression. The generator can synthesize an arbitrary amount of facial identities with different head poses, expressions, backgrounds and illuminations. In the following, we describe the most important parameters of the model and their influence on the facial appearance in the image.
	
	\textbf{Facial identity.} We assume that the facial identity is fully determined by the 3D face shape and color. We use the Basel Face Model 2017 \cite{bfm17}, for which the shape and color distribution is estimated from 200 neutral high-resolution 3D face scans. 
	The parameters follow a Gaussian distribution. By drawing random samples from this distribution we generate random 3D face meshes with unique color and shape.
	
	\textbf{Illumination.} In our renderer, we assume the Lambertian reflectance model. We approximate the environment map with 27 spherical harmonics coefficients (9 for each color channel).  Sampling these coefficients randomly would lead to possibly unrealistic illumination conditions. In order to achieve more natural illumination conditions in the synthetic face images, we sample the spherical harmonics illumination parameters from the Basel Illumination Prior (BIP) \cite{illuprior}. The BIP describes an empirical distribution of spherical harmonics coefficients estimated from 14'348 real-world face images.
	
	\textbf{Pose and Camera}. The face is viewed by a fixed pinhole camera. The face orientation with respect to the camera is controlled by the six pose parameters (yaw, pitch, roll, and 3D translation). Throughout our experiments, we vary the head pose angles in order to simulate different head poses. We normalize the head position to be centered in the image frame.
	
	\textbf{Background}. We simulate changes in the background with a non-parametric background model by sampling randomly from a set of background images of the describable texture database \cite{textures}. The purpose of these random structured background changes is to help the deep learning system in discovering the irrelevance of background structures for the task of face recognition.
	
	The synthesized images are fully specified by the aforementioned parameter distributions which were lear-ned from a population of 3D face scans. Note that the data generator enables the statistical variation of face shapes and textures which is in contrast to e.g. the data augmentation in \cite{masi2016we} where shape deformation between a few fixed 3D shapes is performed.	Another benefit of the generator is that by sampling from its parameters we can synthesize an arbitrary amount of face images with an arbitrary number of identities in different head poses, with expressions and natural illumination conditions.
	
    \textbf{Software}. The face image generator was created with the scalismo-faces library \footnote{https://github.com/unibas-gravis/scalismo-faces/} and is publicly available \footnote{https://github.com/unibas-gravis/parametric-face-image-generator}. It is based on our previous work \cite{kortylewski2017empirically} where we used a simple point light source and a discrete sampling of the parameters to study the influence of nuisance transformations on face recognition systems in the virtual domain. For, this work we extended the generator with an illumination prior, facial expressions and a random sampler for the parameters.
  
	\begin{figure*}[ht]
    \centering
    \subfloat[\label{fig:gapmul}]{\includegraphics[width=.32\textwidth]{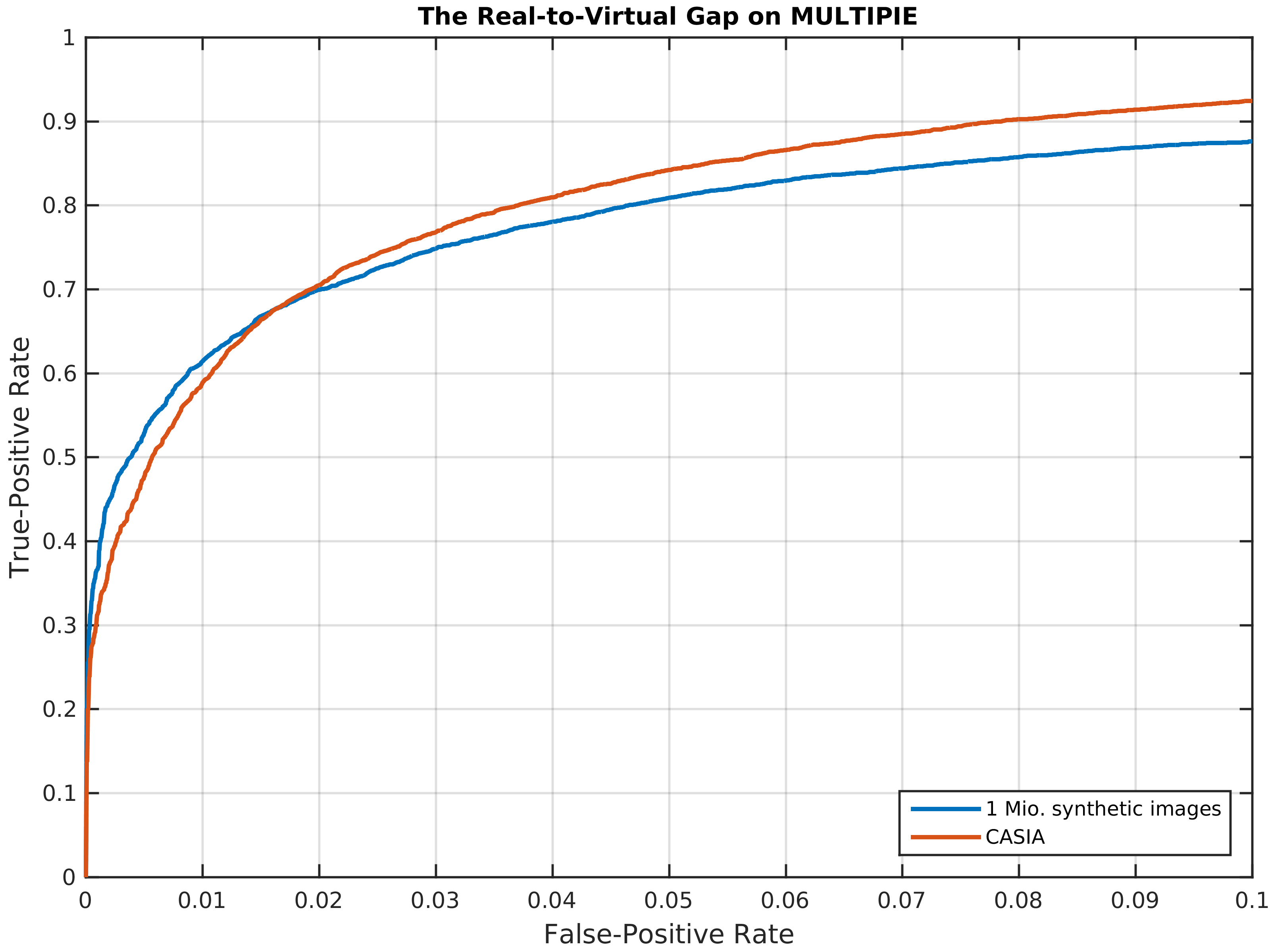}}\quad
    \subfloat[\label{fig:gaplfw}]{\includegraphics[width=.32\textwidth]{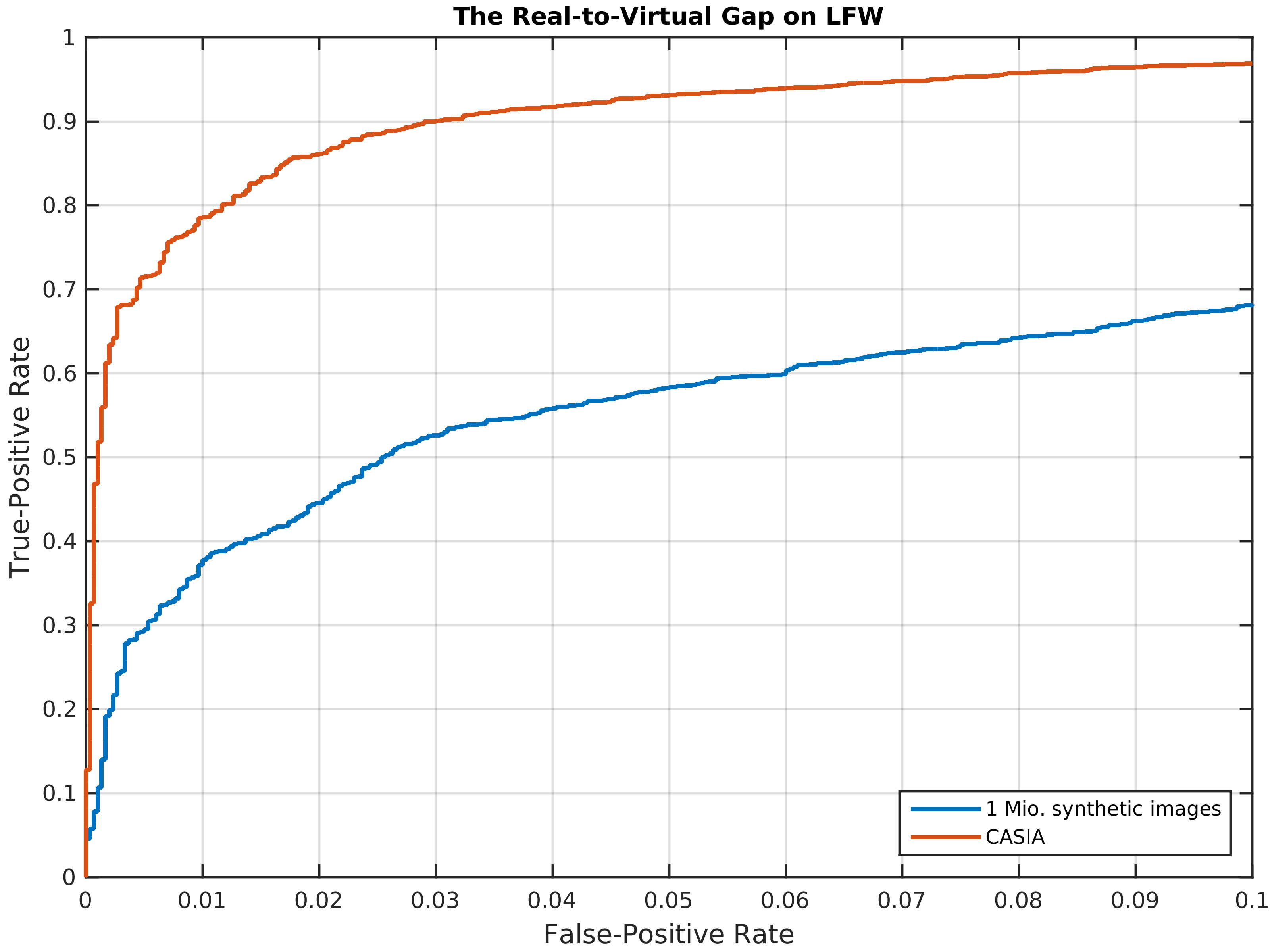}}\quad
    \subfloat[\label{fig:gapijb}]{\includegraphics[width=.32\textwidth]{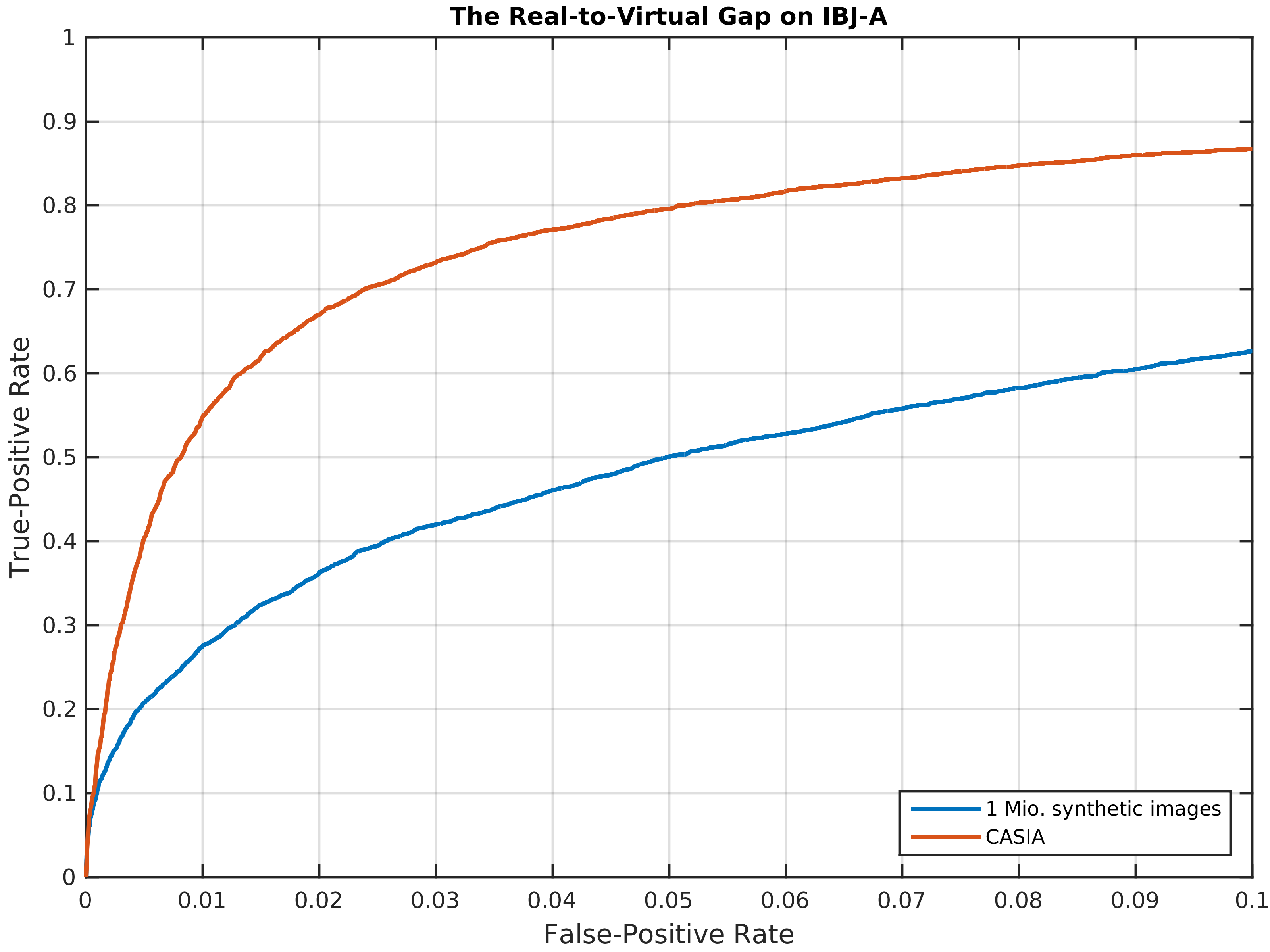}}
    \caption{The real-to-virtual performance gap in face recognition. We compare the ROC curves of the FaceNet-NN4 architecture trained on one million synthetically generated images (blue line) and on the real-world Casia dataset (red line) for three different benchmarks: (a) CMU-Multipie; (b) LFW; (c) IJB-A. On the CMU-Multipie both networks perform similarly, whereas on the complex LFW and IJB-A benchmarks a significant performance gap exists between the to networks.}
    \label{fig:gap}
    \end{figure*}
    \section{Experiments}
    \label{sec:exp}
    In this Section, we study the core research question that guides our article, namely:
    
	\begin{center}
	    \textit{Can we use synthetically generated data to support the training and reliability of face recognition systems?}
	\end{center}  
    
    After introducing our experimental setup (Section \ref{sec:setup}), we study if synthetic data can fully replace additional real data when training face recognition systems (Section \ref{sec:gap}). We show that this hypothesis is indeed valid for the simple CMU-Multipie benchmark. Whereas, for complex face recognition benchmarks such as LFW and IJB-A a significant real-to-virtual gap exists. In Section \ref{sec:gapclose}, we demonstrate that this real-to-virtual gap can be closed by fine-tuning a model trained on synthetic data with real data. 
    Finally, we then leverage the parametric nature of our data generator to explore how changing the characteristics of the synthetic dataset in terms of the pose distribution and the number of identities affects the overall recognition performance (Section \ref{sec:changingSynthData}). 
    \subsection{Experimental Setup}
    \label{sec:setup}
    All our experiments are based on the OpenFace framework \cite{openface}, as the software is publicly available and well documented. For the face detection and alignment we use an implementation of the Multi-task CNN \footnote{https://github.com/kpzhang93/mtcnn\_face\_detection\\ \_alignment} \cite{mtcnn}. To the best of our knowledge, this combination of software is one of the best publicly available face recognition environments that enable the self-training of networks.
    
    We train the FaceNet-NN4 architecture which was originally proposed by Schroff et al. \cite{schroff2015facenet} with the vanilla setting as provided in the OpenFace framework. The aligned images are re-sized to $96\times96$ pixels. The triplet loss is trained with batches of $20$ identities with $15$ sample images per identity for $200$ epochs. In order to keep our measurements unbiased, we do not perform data-augmentation nor dataset adaptation. 
    
    \textbf{Real-world training and benchmark data.} Whenever we train a network with real-world data, the data is sampled from the cleaned Casia WebFace dataset \cite{casia}, which comprises 455,594 images of 10,575 different identities. From this dataset, we remove the 27 identities which overlap with the IJB-A dataset.
    
    For benchmarking we use three popular benchmarks: 1) CMU-Multipie \cite{multipie} was recorded under controlled illumination and with the same background. The head pose of the subjects is uniformly distributed over the full $180^\circ$ yaw range spaced $15^\circ$ apart. We use the identities from session one with the frontal illumination setting. Images from the two overhead cameras are excluded.  2) LFW \cite{lfw} has been the de facto standard face recognition benchmark for many years. Face images in this dataset are subject to a complex illumination, partial occlusion, and background clutter. However, due to its strong bias towards frontal poses \cite{yin2017multi} modern deep face recognition systems achieve close to perfect results on LFW. 3) IJB-A \cite{ijba} was proposed to further push the frontiers of face recognition. The conditions regarding pose, illumination and partial occlusion are more complex compared to LFW. Another interesting property is that the subjects might be described by multiple gallery images. These image sets are commonly referred to as \textit{templates}. If the Multi-task CNN face detection fails to detect a face, we use hand cropped face boxes. For LFW and IJB-A these crops are provided with the datasets. For Multipie, we use the annotations provided by El Shafey et al. \cite{el2013scalable}.
    
    \textbf{Synthetic face image generation setup.} The synthetic face images are generated by randomly sampling the parameters of our face image generator (Section \ref{sec:generator}). Our synthetic dataset consists of 1 Million images with $10K$ different identities and $100$ example images per identity. Thereby, the shape, color and expression parameters are sampled according to the Gaussian distributions from the Basel Face Model \cite{bfm17}. The parameters for the illumination are randomly sampled from the empirical Basel Illumination Prior \cite{illuprior}. The head pose is randomly sampled according to a uniform pose distribution on the yaw, pitch and roll angles in the respective ranges $r_{yaw}=[ -90^\circ,90^\circ ]$, $r_{pitch}=[ -30^\circ,30^\circ ]$ and $r_{roll}=[ -15^\circ,15^\circ ]$. We have chosen this pose distribution to achieve a good generalization across the full variation of the head pose in facial images. This choice is contrasting most of the web crawled datasets, which have biases towards frontal yaw poses. For each synthetic image, we randomly select a background texture, as explained in Section \ref{sec:generator}. 
    Throughout the rest of this article, we refer to this dataset as \textit{SYN-1M}.
    
    \noindent\textbf{Evaluation protocol.} We evaluate face recognition networks at the task of face verification. Thereby, we measure the distance between two face images as the cosine distance between their 128-dimensional feature embeddings from the last layer of the FaceNet model:
    \begin{equation}
        s(a,b)=\frac{\sum_i a_i b_i}{\sqrt{\sum_i a_i^2}\sqrt{\sum_i b_i^2}} \hspace{0.2cm}.
    \end{equation}
   For comparing the templates in the IJB-A dataset, we perform softmax averaging of the similarity scores between each image pair as proposed by Masi et al. \cite{masi2016we}. We do not perform any dataset adaptation, thus we test the most challenging face recognition setup with only \textit{unrestricted, labeled outside data}. For LFW and IJB-A, the pairwise comparisons are given by the respective protocols. For the Multipie dataset, we follow the LFW protocol. Thus, we perform 10 fold cross validation with 300 pairs of positive and negative samples.    
    \begin{figure*}
    \centering
    \subfloat[\label{fig:gapclose-mul}]{\includegraphics[width=.3\textwidth]{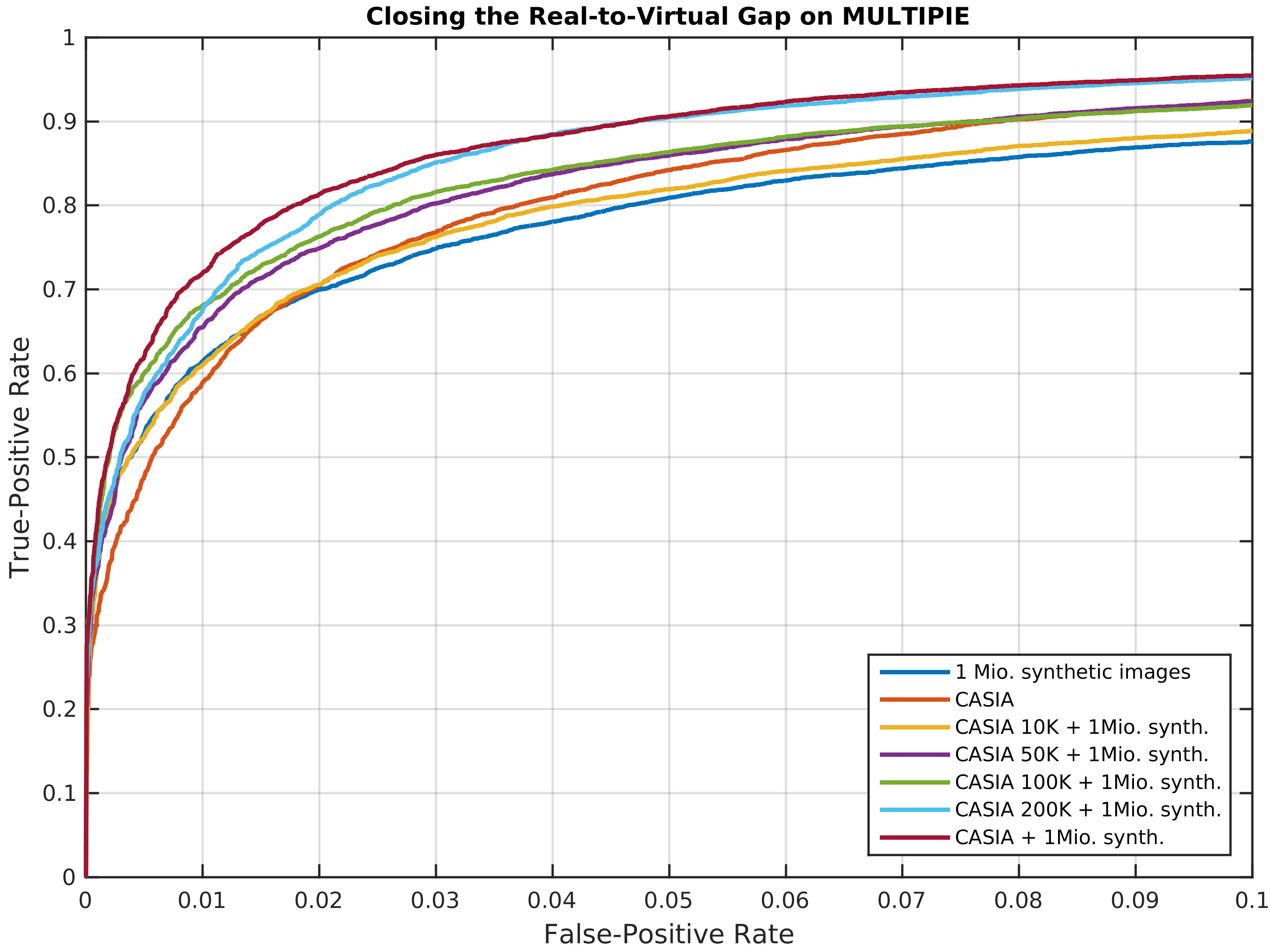}}\quad
    \subfloat[\label{fig:gapclose-lfw}]{\includegraphics[width=.3\textwidth]{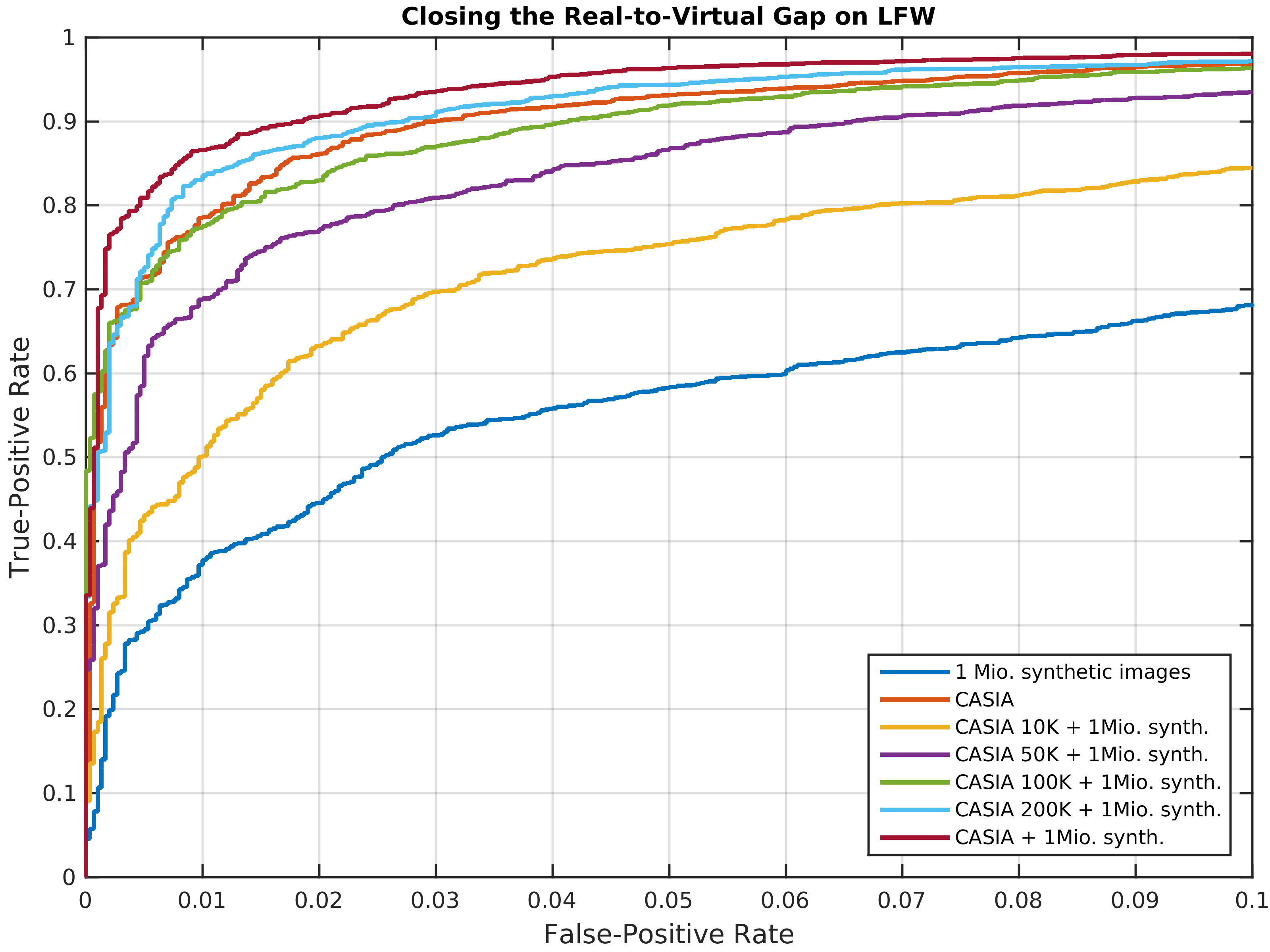}}\quad
    \subfloat[\label{fig:gapclose-ijb}]{\includegraphics[width=.3\textwidth]{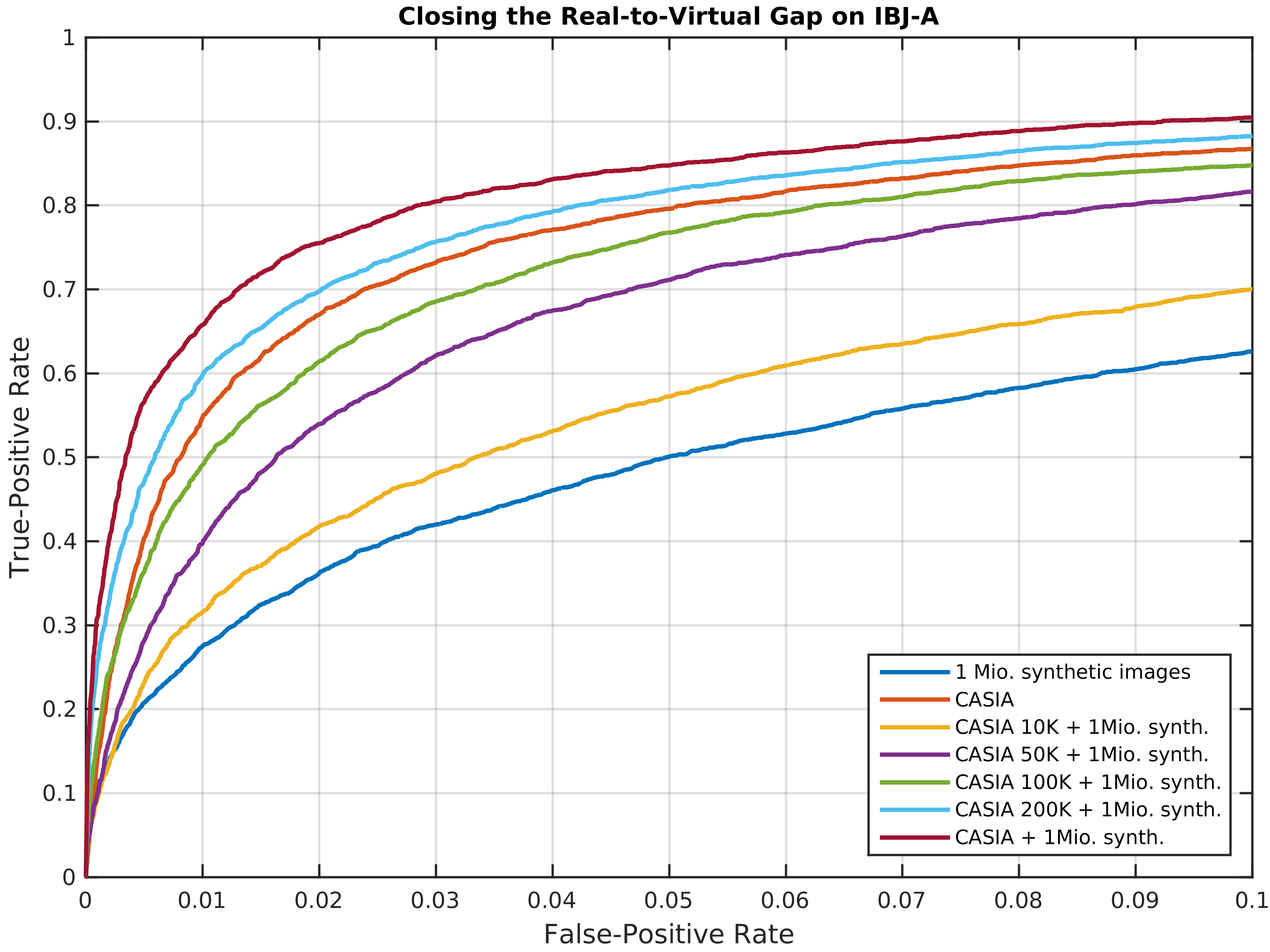}}
    \caption{Closing the real-to-virtual gap by fine-tuning on real-world data. We fine-tune FaceNet-Synth with different amounts of real-world data ($N=\{10K ,50K ,100K ,200K ,453K\}$) and compare the respective ROC curves on: (a) CMU-Multipie; (b) LFW; (c) IJB-A. The prominent real-to-virtual gap on LFW and IJB-A can be closed with about $100K-200K$ images. Fine-tuning with the full $453K$ Casia dataset leads to a significantly improved performance across all three benchmarks compared to training without synthetic data.}
    \label{fig:gapclose}
    \end{figure*}

    \subsection{The real-to-virtual gap in face recognition}
    \label{sec:gap}
    In the following experiments, we measure the recognition performance of a deep face recognition system trained on fully synthetic data and compare it to a system trained on real-world data. Thereby, we train a FaceNet-NN4 model with the Casia dataset (referred to as FaceNet-Casia) and the synthetically generated SYN-1M data (referred to as FaceNet-Synth). The performance of the trained networks is illustrated in Figure \ref{fig:gap}. On the CMU-Multipie benchmark (Fig. \ref{fig:gapmul}), both networks perform similarly, suggesting that 3DMMs can very well model the facial appearance variation in a controlled setting. Whereas in the more challenging settings of LFW (Fig. \ref{fig:gaplfw}) and IJB-A (Fig. \ref{fig:gapijb}) the FaceNet-Synth model performs significantly worse than FaceNet-Casia. Hence, \textit{a prominent real-to-virtual gap can be observed}. To the best of our knowledge, this is the first measurement of this gap at the task of face recognition. Our measurements contrast the findings of Gaidon et al. \cite{gaidon2016virtual} who have shown that deep learning systems for multi-object tracking trained on synthetic data perform practically as well as if they were trained on real data. This discrepancy may arise due to the property that for object tracking rather high-level features are important which can be simulated well. In contrast, faces differ in comparatively subtle small-scale features which are so far not possible to simulate with a fully parametric model. In addition, the Basel Face Model, which our generator is based on, has a bias towards rather young Caucasian people and does not model facial hair or partial occlusion. Overall, we suppose that this model mismatch induces the observed real-to-virtual performance gap.
    In the next Section, we will explore if the real-to-virtual gap can be overcome by combining real and synthetic data.
    \subsection{Closing the real-to-virtual gap}
    \label{sec:gapclose}
    In the following series of experiments, we study how the real-to-virtual gap between synthetically generated images and the face recognition benchmarks of LFW and IJB-A can be closed. Following the commonly applied scheme for transfer learning with deep neural networks, we fine-tune the FaceNet-Synth model with real-world data. In order to study the influence of the dataset size, we randomly sample subsets of the Casia dataset with size $N=\{10K,50K,100K,200K\}$. We also test the effect of using the full Casia dataset of about $453K$ images for fine-tuning. Figure \ref{fig:gapclose} compares the performance of the fine-tuned networks with FaceNet-Synth and FaceNet-Casia over all our benchmark datasets.
    
    \textbf{General observations.} We can observe the following general effects: 1) Fine-tuning FaceNet-Synth with any amount of real data leads to an increase of the recognition performance compared to the original FaceNet-Synth and thus reduces the real-to-virtual gap. 2) The size of the realistic dataset used for fine-tuning positively correlates with the final recognition performance. 3) With a subset of only $100-200K$ images, the real-to-virtual gap is closed and the fine-tuned network outperforms FaceNet-Casia on all benchmarks. 4) Using the full Casia dataset for fine-tuning even increases the performance gain.
    
    \textbf{Benchmark-specific observations:} On the LFW dataset (Figure \ref{fig:gapclose-lfw}) the real-to-virtual gap can be almost closed with $100K$ realistic training examples. Thus, pre-training on synthetic data reduced the amount of data needed to achieve similar performance as FaceNet-Casia by about a factor of five. We are not aware of any related work that achieves similar results with such a small amount of real data without performing any data augmentation nor any dataset adaptation. For the IJB-A dataset (Figure \ref{fig:gapclose-ijb}), $100K$ images do not suffice to close the real-to-virtual gap. This can be attributed to the additional complexity of IJB-A compared to LFW. When fine-tuning with $200K$ images, we achieve a significantly improved performance compared to the FaceNet-Casia network. Thus, on the IJB-A dataset, the amount of data needed to achieve competitive performance is more than halved. On the Multipie dataset (Fig. \ref{fig:gapclose-mul}), fine-tuning with even very small amounts of data, such as e.g. $10K$ images, leads to an improved performance compared to FaceNet-Casia.
    
    Table \ref{tab:perf} summarizes the results of Figure \ref{fig:gapclose} and compares the performance of fine-tuned FaceNet-Synth networks to those models that are solely trained on the real-world data subsets. Pre-training on synthetic data always increases the performance compared to training with real data only. For the small $50K$ subset the performance increase is very prominent and already results in highly competitive performances. Notably, pre-training on synthetic data leads to a performance increase across all benchmarks, even though the benchmarks have very different imaging characteristics.
    
    Overall, the results in Figure \ref{fig:gapclose} and Table \ref{tab:perf} show that the real-to-virtual gap can be closed by fine-tuning on relatively small amounts of real data. Using the full Casia dataset to fine-tune FaceNet-Synth significantly increases the recognition performance across all datasets compared to using the real data only. Thus, suggesting that knowledge can be transferred from the virtual to the real domain. The additional recognition performance is likely to be induced by the complementary characteristics of the Casia and the SYN-1M dataset. SYN-1M has more samples per identity than Casia and has a much broader pose distribution. Whereas, images in Casia show facial properties which, so far, cannot be modeled with 3DMMs such as facial occluders or hair.
    
    In the next Section \ref{sec:changingSynthData}, we will leverage the parametric nature of the synthetic data generator to study the influence of the pose-distribution and the number of identities in the synthetic dataset on the recognition performance.
    
    \begin{table}
	    \centering
		\begin{tabular}{l|c|c|c}
			\hline
			Datasets & \textbf{Multipie} & \textbf{LFW}  & \textbf{IJB-A} \\
			\hline  
			Metric              & Accuracy  & Accuracy  & TAR \\ 
			\hline  
			SYN-1M              & 0.889     & 0.801      & 0.625 \\
		\hline\hline  
			Casia-10            & 0.759     & 0.750      & 0.407  \\ 
			\hline  
			Casia-10 + SYN-1M   & 0.895      & 0.827    & 0.701 \\
			\hline\hline
			Casia-50            & 0.817     & 0.851     & 0.662 \\ 
			\hline
			Casia-50 + SYN-1M   & 0.913      & 0.918    & 0.834 \\
			\hline  \hline
			Casia-100           & 0.836     & 0.891     & 0.713 \\ 
			\hline 
			Casia-100 + SYN-1M  & 0.913    & 0.936     & 0.850 \\
			\hline\hline  			
			Casia-200           & 0.860     &  0.927   & 0.819 \\ 
			\hline
			Casia-200 + SYN-1M  & 0.930    & 0.948     & 0.882 \\ 
			\hline\hline  			
		Casia               & 0.912      & 0.941     & 0.868 \\ 
			\hline  
			Casia + SYN-1M      & \textbf{0.933}     & \textbf{0.958}     & \textbf{0.906} \\ 
			\hline\hline  
		\end{tabular}
		\caption{Performance comparison between networks pre-trained on synthetic data and fine-tuned on real data against networks trained solely on real data across the CMU-Multipie, LFW and IJB-A benchmarks. We test different sized subsets $\{10K,50K,100K,200K,453K\}$ from the Casia dataset. We measure the recognition accuracy and the $TAR$ at $FAR=0.1$. For all sizes of the real dataset, fine-tuning on synthetic data induces significant performance improvements across all benchmarks.}
		\label{tab:perf}
	\end{table}
	    \begin{table}
	    \centering
	    \subfloat[Original SYN-1M performance]{    
			\begin{tabular}{l|c|c|c}
				\hline
				Datasets & \textbf{Multipie} & \textbf{LFW}  & \textbf{IJB-A} \\
				\hline  
				Metric                  & Accuracy  & Accuracy  & TAR \\ 
				\hline  
				SYN-1M                  & 0.893     & 0.801      & 0.625 \\
			    \hline  
			    Casia + SYN-1M \hspace{.175cm}     & 0.933     & 0.958     & 0.906 \\
				\hline\hline  
			\end{tabular}
	    }\\	
	    \subfloat[Bias to frontal pose]{
			\begin{tabular}{l|c|c|c}
				\hline
				Datasets & \textbf{Multipie} & \textbf{LFW}  & \textbf{IJB-A} \\
				\hline  
				Metric                  & Accuracy  & Accuracy  & TAR \\ 
				\hline  		
				SYN-Front            & 0.755     & 0.706      & 0.405 \\
				\hline  
				Casia + SYN-Front    & 0.911   & 0.931      & 0.827 \\ 
				\hline\hline  
			\end{tabular}
	    }\\		
	    \subfloat[Double the number of identities]{
			\begin{tabular}{l|c|c|c}
				\hline
				Datasets & \textbf{Multipie} & \textbf{LFW}  & \textbf{IJB-A} \\
				\hline  
				Metric                  & Accuracy  & Accuracy  & TAR \\ 
				\hline  
				SYN-2M                  & 0.899     & 0.771      & 0.546 \\
				\hline  
				Casia + SYN-2M \hspace{.175cm}         & 0.954      & 0.960    & 0.924 \\
				\hline\hline
			\end{tabular}
		}	
			\caption{Effect of changing the characteristics of the synthetic dataset on the recognition performance. We compare the performance of neural networks trained on: (a) The original SYN-1M dataset. (b) A synthetic dataset that has a strong bias toward frontal yaw pose ($r_{yaw}=[ -35^\circ,35^\circ ]$). (c) A synthetic data set with two million images that was generated by doubling the amount of identities to $20K$. We measure the recognition accuracy and the $TAR$ at $FAR=0.1$. Biasing the synthetic data to frontal poses significantly reduces the recognition performance. Doubling the number of identities leads to an increased performance.}
			\label{tab:character}
		\end{table}
    \subsection{Changing the characteristics of the synthetic dataset}
    \label{sec:changingSynthData}
    In this section, we demonstrate how the fully parametric nature of the face image generator can be used to study important characteristics of large-scale face image datasets. In the following, we study the effect of the characteristics of the head pose distribution and the number of identities in the synthetic dataset on the recognition performance. Importantly we change one dataset characteristic at a time while keeping all other parameters of our data generator fixed. The experimental results are summarized in Table \ref{tab:character}. 
    
    \textbf{Bias to frontal pose.} We generate the SYN-1M-Front dataset which, compared to the SYN-1M dataset, is limited in the yaw angle to the range of $r_{yaw}=[ -35^\circ,35^\circ ]$. With this setup, we simulate a bias towards frontal head poses, which is prevalent in many datasets. Training the FaceNet-NN4 architecture with the SYN-1M-Front data induces a significant performance decrease compared the SYN-1M dataset, in which the pose distribution varies across the full yaw angle (Table \ref{tab:character}). The performance decrease is also present after fine-tuning on the Casia dataset. These results demonstrate that the additional knowledge about pose variation in the SYN-1M dataset induces an improved performance on the real-world benchmarks. Thus, proving that deep neural networks are capable of conserving and transferring knowledge about the virtual world to the real-world when performing face recognition.
    
    \textbf{Increasing the number of training identities.} In this setup, we double the number of identities to $20K$, thus generating two million synthetic training images. Note that the collection of one million additional images requires an enormous effort for a real-world dataset, whereas for synthetic data it can be done effortlessly. The network trained solely on the SYN-2M dataset performs slightly worse than with the SYN-1M data, which might be due to some overfitting to the synthetic domain. However, after fine-tuning an increase in the recognition performance can be observed (Table \ref{tab:character}). The additional benefit in terms of recognition performance is, however, relatively small when taking into account that the amount of data has doubled. This is in line with the observation made by Schroff et al. \cite{schroff2015facenet} who observed this phenomenon with real data.

    \section{Conclusion}
    \label{sec:conclusion}
    This paper makes several important contributions to training deep face recognition systems with synthetic data.
    
	First of all, we document a significant gap between the face recognition performance of neural networks trained with real data compared to networks trained with synthetic data, which is generated by a state-of-the-art synthetic face image generator. This observation was to be expected as the face generator is not able to simulate facial appearance details, facial occluders, and partial occlusion.
	
	Second, we showed that by using synthetic data in combination with real-world data, large performance gains can be achieved with a significantly reduced amount of real-world data. We explain this observation by the fact that the face image generator is capable of simulating strong pose variation and the resulting self-occlusion, as well as statistical variations of the face shape, albedo, and illumination. In a real-world dataset, it is difficult to cover the full range of these variations.
    Our observations demonstrate that knowledge transfer between the virtual and the real-world data is highly beneficial for the performance of face recognition systems.
    
    Third, we demonstrated that the additional variability of the synthetic data in terms of pose and facial identity are crucial factors for the increased performance on real-world data. Our experiments highlight the strong benefits of a fully parametric data generator, which makes possible to effortlessly adjust the characteristics of the dataset and thus to perform quantitative studies of dataset properties, which are very difficult to control with real-world data.

	Finally, in order to support the reproducibility of our results and any future developments, we make the data generator publicly available. 
    Our software makes possible to generate large face datasets with complex annotations in a less time-consuming manner.
    
    Our work opens several future research directions with a potentially major impact. The proposed approach can generally be applied to other computer vision domains which use statistical object models. 
    Especially in the field of medical image analysis, where data is often a limiting factor, synthetic data generation and augmentation using statistical shape models would offer an alternative approach.	
	Finally, improving the realism of the face image generator could further decrease the amount of real-world data needed for training competitive deep face recognition systems. Ultimately, this approach even has the potential to alleviate one of the currently most challenging problems in facial image analysis which is the data collection process.\\

	\section*{Acknowledgment} We gratefully acknowledge the support of NVIDIA with the donation of a Titan Xp and a Quadro P5000.

	\bibliographystyle{spmpsci}
	\bibliography{egbib}  
	
\end{document}